\documentclass[conference]{IEEEtran}
\usepackage{times}

\usepackage[numbers]{natbib}
\usepackage{booktabs} %
\usepackage[ruled,vlined]{algorithm2e}
\usepackage{algpseudocode}

\usepackage{multicol}
\usepackage[bookmarks=true]{hyperref}
\usepackage{graphicx}
\usepackage{amsmath}
\usepackage{cleveref}
\usepackage{xcolor}
\newcommand{\numberthis}[1]{\addtocounter{equation}{1}\tag{\theequation}\label{#1}}
\DeclareMathOperator*{\argmax}{arg\,max}
\DeclareMathOperator*{\argmin}{arg\,min}

\usepackage{xcolor}

\definecolor{burntorange}{rgb}{0.8, 0.33, 0.0}
\usepackage{color, colortbl}
\definecolor{LightCyan}{rgb}{0.88,1,1}
\definecolor{Gray}{gray}{0.9}

\newcommand{\statspace}{\mathcal{S}}
\newcommand{\obsspace}{\Omega}
\newcommand{\actspace}{\mathcal{A}}
\newcommand{\transfn}{T}

\newcommand{\task}{\mathcal{T}}
\newcommand{\langfeedback}{\mathcal{L}}
\newcommand{\state}{s}
\newcommand{\action}{a}
\newcommand{\obs}{o}
\newcommand{\costbasic}{\mathcal{C}_{\mathcal{B}}}
\newcommand{\planner}{\mathcal{P}}
\newcommand{\cmap}{\mathcal{C}}

\newcommand{\env}{\mathcal{E}}
\newcommand{\humano}{o^h}

\newcommand{\costhuman}{\mathcal{C}_{\mathcal{H}}}

\newcommand{\robotcost}{\mathcal{C}_{\mathcal{R}}}
\newcommand{\starrobotcost}{\mathcal{C}^*_{\mathcal{R}}}

\newcommand{\taskcost}{\mathcal{C}_{\mathcal{T}}}
\newcommand{\languagecost}{\mathcal{C}_{\mathcal{L}}}
\newcommand{\mask}{\mathcal{M}}

\newcommand{\ddemo}{\mathcal{D}_{\text{demos}}}
\newcommand{\dcmap}{\mathcal{D}_{\text{cmap}}}

\usepackage{amssymb}%
\usepackage{pifont}%

\usepackage{multirow, makecell}

\usepackage[font=small,labelfont=bf]{caption}
\title{\huge Correcting Robot Plans with Natural Language Feedback}

\author{
\authorblockN{
Pratyusha Sharma\authorrefmark{3}\authorrefmark{4},
Balakumar Sundaralingam\authorrefmark{3},
Valts Blukis\authorrefmark{3},
Chris Paxton\authorrefmark{3},\\
Tucker Hermans\authorrefmark{3}\authorrefmark{5},
Antonio Torralba\authorrefmark{4}, %
Jacob Andreas \authorrefmark{4},
Dieter Fox\authorrefmark{3}\authorrefmark{2}}\vspace{0.1cm}
\authorblockA{\authorrefmark{3} NVIDIA,
\authorrefmark{4} MIT,
\authorrefmark{5} University of Utah,
\authorrefmark{2} University of Washington}
}
\begin{document}

\setcounter{figure}{1}
\makeatletter
\let\@oldmaketitle\@maketitle%
\renewcommand{\@maketitle}{\@oldmaketitle%
\begin{center}
    \centering     
    \includegraphics[width=\textwidth,]
    {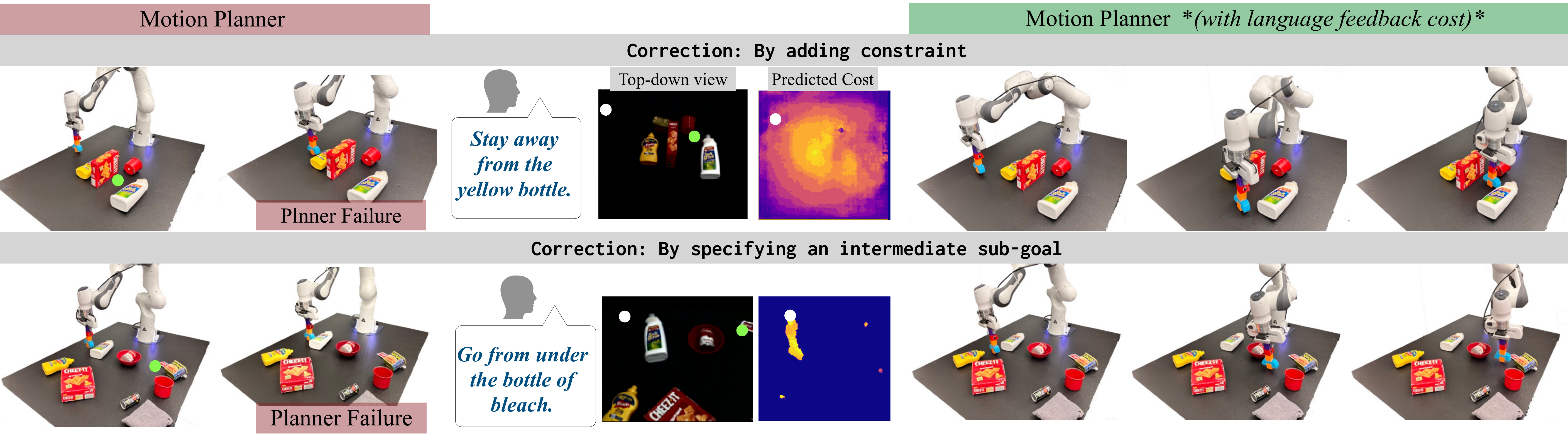}
  \end{center}
  \footnotesize{\textbf{Fig.~\thefigure:\label{fig:intro}}~ Robots often fail to do what we want. This can happen for many reasons including mis-specification of goals, failure to anticipate what satisfying plans will do, and because optimization sometimes fails. We show how language can be used to update the underlying cost of a planner to improve task performance.
  Our approach can use language to specify corrections by a) the addition of constraints or b) specifying intermediate sub-goals for the planner. 
  }\vspace{-12pt}
  \medskip}%
\makeatother
\maketitle

\begin{abstract}

When humans design cost or goal specifications for robots, they often produce specifications that are ambiguous, under-specified, or beyond planners' ability to solve. In these cases, \emph{corrections} provide a valuable tool for human-in-the-loop robot control.
Corrections might take the form of new goal specifications, new constraints (e.g.\ to avoid specific objects), or hints for planning algorithms (e.g.\ to visit specific waypoints).  Existing correction methods (e.g.\ using a joystick or direct manipulation of an end effector) require full teleoperation or real-time interaction.
In this paper, we explore \emph{natural language} as an expressive and flexible tool for robot correction.
We describe how to map from natural language sentences to \emph{transformations of cost functions}. We show that these transformations enable users to correct goals, update robot motions to accommodate additional user preferences, and recover from planning errors.
These corrections can be leveraged to get 81\% and 93\% success rates on tasks where the original planner failed,
with either one or two language corrections.
Our method 
makes it possible to compose multiple constraints and
generalizes to unseen scenes, objects, and sentences in simulated and real-world environments. Additional visualizations are available at~\href{https://sites.google.com/view/language-costs}{sites.google.com/view/language-costs}

\end{abstract}

\section{Introduction}
\label{sec:introduction}

Consider a robot vacuum cleaner. The robot's goal is to clean the house, but there may be a need to alter the objective (\emph{``Clean only the living room.''}), to introduce constraints (\emph{``Don’t go into the bathrooms!''}) or to guide the robot when it is stuck  (\emph{``Go to the right end of the wall to enter the missed room.''}). The robot would benefit from the ability to incorporate such corrective, natural language feedback to alter aspects of its behavior or modify its goal. How though can the robot incorporate instructions with rich and varied semantics into its existing objective?

In this paper, we propose to use natural language instructions as inputs to directly modify a robot's planning objective. This objective function takes the form of a cost function in an optimization-based planning and control framework for manipulation.
Our use of language contrasts with previous work where corrective input of robot behavior came from joystick control~\cite{Spencer-RSS-20,rakita2018autonomous}, kinesthetic feedback~\cite{Muxfeldt2014KinestheticTI, Jain2015LearningPF, bajcsy-hri2018}, or spatial labelling of constraints~\cite{xia2013model,bowyer2013active}.
Kinesthetic and joystick feedback allows for fine-grained control, but typically requires prior expertise and undivided attention from the user, reducing the system autonomy and limiting its applications.
\begin{figure}[ht]
    \centering
    \includegraphics[width=0.49\textwidth]{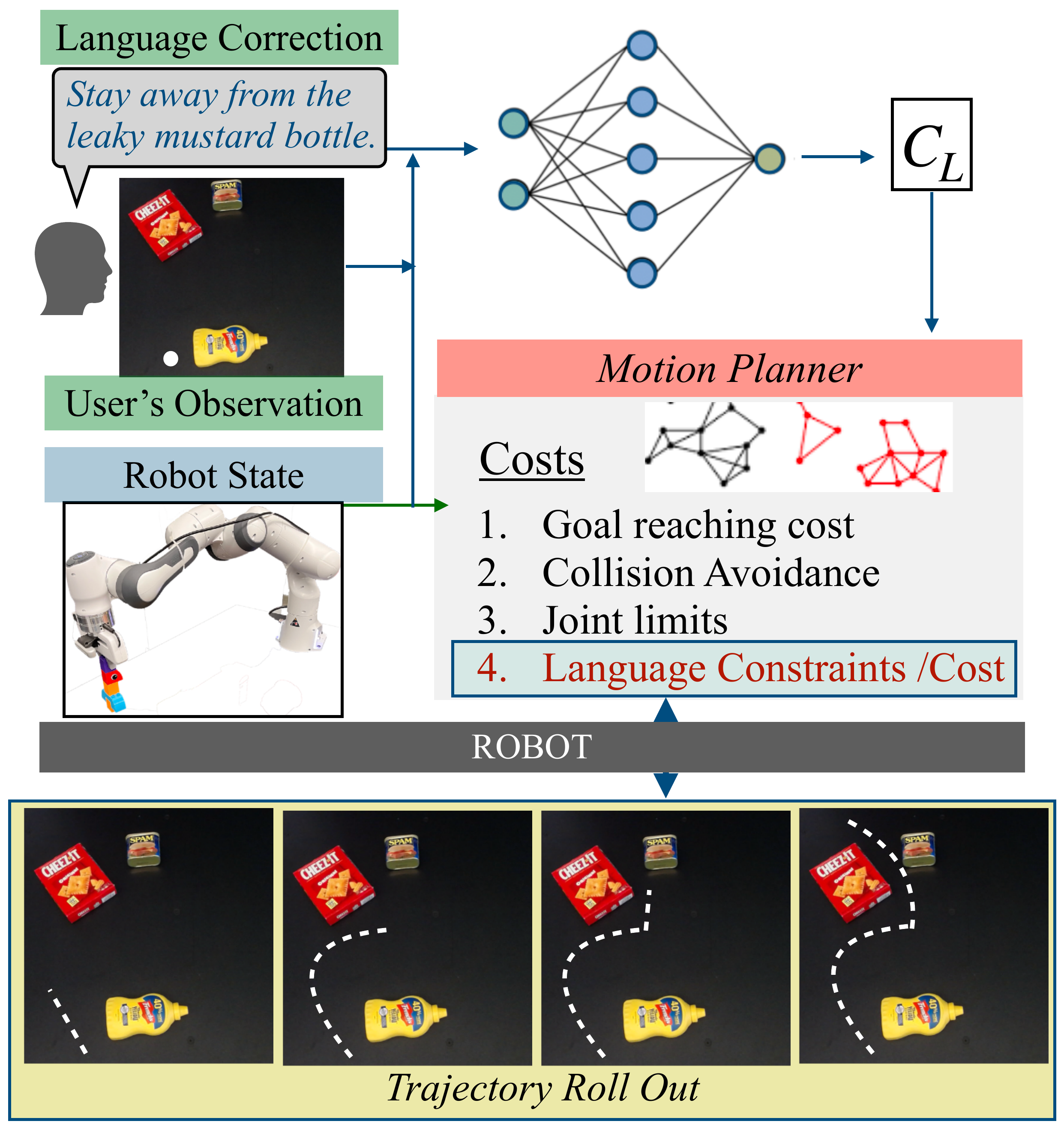}
    \caption{\textbf{Setup:} The main components of our system are: the motion planner and language parameterized cost correction module. The motion planner uses sampling based model predictive control to minimize the overall specified cost. The cost correction module takes as input its observation of the environment, the robot's state, and a natural language correction to output an updated cost function. The motion planner uses the updated cost to modify the trajectory of the robot.}
    \label{fig:framework}
\end{figure}

We choose natural language input thanks to its efficiency, accessibility, ease-of-use, and direct conveyance of the user's intent~\cite{tellex2020-robots-that-use-language}. This allows for the expression of a broad range of feedback describing physical variation, time, or other abstractions.
However, language also brings with it ambiguity and requires a model of symbol grounding and spatio-temporal reasoning to relate the concepts expressed in language to the robot's state and action spaces~\cite{Harnad1990SymbolGP, Mooney2008LearningTC, Tellex2011UnderstandingNL, Matuszek2012LearningTP, Andreas2015AlignmentBC, Paul2016EfficientOA, Padmakumar2018LearningAP}.
Given such a grounding model, the robot can ground language corrections to novel tasks and environments, achieving powerful generalization.
In contrast, learning to generalize corrections from kinesthetic or joystick data to novel tasks requires inferring the user's intent given few underspecified demonstrations, which is itself a challenging problem~\cite{Dragan2013APF}.

We propose to learn a model that maps visual observations and a natural language correction to a residual cost function.
Specifically, we model natural-language corrections as a residual cost function 
that 
can be combined with a task cost. %
This allows a user to modify the robot's objective, to clarify a misspecified objective, or introduce additional constraints into the motion optimization process at varying levels of abstraction at any time during execution.
In the absence of a prior task objective, our method can also be used to specify the task in an instruction-following setting~\cite{misra_tellmedave_2014,venkatesh-iros2021}.
Our framework seamlessly integrates with commonly used motion planner costs, such as collision avoidance, joint limits, and smoothness.
It also allows layering costs sequentially or at a given time, allowing for time-varying corrections.
Finally, it enables composing costs associated with previously learned tasks or corrections to represent new tasks at a higher level of abstraction.
We train our cost model on a dataset of natural language descriptions paired with either demonstrations, pre-specified costs, or both.

We conduct experiments both in simulation and on the physical robot manipulator illustrated in Figure~1.
These experiments show how we can use our method to either specify or correct robot behavior with natural language commands. 
In environments where our local planner fails $6.4\%$ of the time, humans can use our language interface to correct $81\%$ of failures with one natural language command and $93\%$ with a second, bringing the effective success rate from $94\%$ to $99\%$. Our method generalizes to unseen objects and to out-of-distribution natural language, and the cost maps it creates are composable, meaning that commands can be combined with various cost functions. Finally, we show how our method, trained in simulation, can be applied to real-robot tasks.

\section{Preliminaries}

Given an environment $\env$, we study planning problems  formalized as Markov decision processes
defined by the tuple $(\statspace,\actspace,\obsspace,\transfn)$. Where $\statspace$ is the set of robot states, $\actspace$ is the set of possible actions, $\obsspace$ is the space of external observations, and $\transfn : \statspace \times \actspace \times \statspace \rightarrow [0,1]$ is the stochastic transition function. In the case of the robot in Figure~\ref{fig:framework}, $\statspace$ is a tuple of all possible positions, velocities and accelerations the robot can achieve represented as $(q,\dot{q},\ddot{q})$.
The action space $\actspace=\{$\emph{up,down,left,right}\(\}\). Its $\obsspace$ is the observed cost of states \(k\)-steps around the robot’s current $\state$. 
The tuple $(\state,\action,\obs,\task)$ describes the robot's state $\state \in \statspace$, action taken $\action \in \actspace$, observations from the environment $\obs \in \obsspace$  and the task $\task$ that the robot needs to complete. For a $\task$, an  associated cost map represented as $\taskcost$ may be specified by a user that wants a planner to perform $\task$. $\taskcost$ is of the form,
\begin{align*}
    \taskcost &:  \statspace \to \mathbb{R} %
\end{align*}
and is a user specified costmap which can be probed for different
$\state \in \statspace$. For example, for reaching a goal, $g$ in
Figure~1 the costmap is a function that at every state
on the map $\state$ returns the Euclidean distance $\sqrt{(s-g)^2}$. %
In addition to the~$\taskcost$, there is also a base cost~$\costbasic$\footnote{Detailed specifications of $\costbasic$ can be found in the \cref{app:mpc}} that helps the robot avoid its limits (collision avoidance, joint limits). 
Conditioned on $\obs$ the robot takes an $\action$ sequentially to minimize,
\begin{align}
    \robotcost=\taskcost+\costbasic.
\end{align}

The planning routine $\planner$ then optimizes the robot's cumulative cost  $\robotcost$ 
and outputs an estimate of the best action that can be taken in the given state $\state$. The robot then takes the action $\action$ and ends up in a new state dictated by the dynamics of the robot $\transfn$ and the system and the processes is repeated. It is via this process that robot unrolls a trajectory, $\tau$, to complete different tasks. More formally, $\planner$ is,

\begin{align*}
    \argmin_{\ddot{q}_t\in[0,T]} \quad & \sum_{t=0}^{T} \taskcost(q_t, \dot{q}_t) + \costbasic(q_t, \dot{q}_t)\\
    \text{s.t.} \quad &  q_t = q_{t-1} + \dot{q}_t dt \numberthis{eq:position} \\ 
    \quad & \dot{q}_t = \dot{q}_{t-1} + \ddot{q}_t dt \numberthis{eq:vel}
\end{align*}

With increasing complexity of the environment and without assuming access to a model of the world, it is challenging to specify
a $\taskcost$ that accurately reflects the task. An optimal cost function would be one that reflects the intended task, capturing the true cost-to-go. For example, the cost function corresponding to navigating to a goal could be approximated as the euclidean distance to goal. However, in many environments, as shown in Fig~1, %
greedily optimizing this objective will result in only a locally optimal solution that is 
completely misaligned with the human's intent. Misalignment due to mis-specified goals and insufficient constraints can cause a plan failure. This problem of misalignment or a difference in the cost in the mind of the human $\costhuman$ and the robot $\robotcost$ is referred to as the value alignment problem \cite{Wallach2008MoralMT}. Feedback from the human can be used to 
minimize the differences between $\costhuman$ and $\robotcost$. The human can generate a feedback based on their observation of the $\env$, represented as $\humano$ and other variables of the $\env$ they have access to.

\section{Approach}
\label{approach}
To minimize the gap between $\costhuman$ and $\robotcost$, we propose the use of feedback from a user in the form of \emph{natural language corrections} to update $\robotcost$. Below we outline our approach.

\subsection{Our Method}
At any given point of time $t$, the user can issue feedback in the form of a natural language string, denoted as $\langfeedback$. We assume that the user has access to $\humano$, $\state$, and information about the task while generating feedback. We learn a generative model that generates a costmap over all $states$ associated with $\langfeedback$ conditioned on $\langfeedback$, $\state$ the state of the robot, and $\humano$.
This cost is then composed with $\robotcost$
($\starrobotcost=\robotcost+\languagecost$) 
to generate an updated cost for the robot. $\planner$ then solves the optimization informed by the updated objective
$\starrobotcost$.
     
We factor the language-based cost~$\languagecost$ into functions that generate a continuous cost map $\cmap$ and a binary mask over the cost $\mask$. We combine them using  element-wise multiplication, \(\languagecost  = \cmap * \mask\). Where the functions have the form,
\begin{align}
    \cmap :&\;  \statspace \to \mathbb{R} \quad \notag\\
    \mask :&\;  \statspace \to [0,1].\notag
\end{align}
$\cmap$ for a given $\langfeedback$ maps to the cost for every $\state \in \statspace$. $\mask$ maps to a binary mask that is used over $\cmap$. 
In the case of a goal-directed $\langfeedback$, such as \emph{go to the left of the bottle}, this is a guiding path to the goal. Whereas in the case of a constraints such as, \emph{go slower}, this is a unit mask i.e. no-states are masked. This is done in-order to use a cue from the mask to help distinguish between changes in goals versus adding constraints to existing goals better. In theory, the cost-map itself should be able to direct a robot to the goal but we see that learning such a decomposition worked better in practice, specially in long-horizon tasks as shown in Appendix~\ref{app:converge}.

$\cmap$ and $\mask$ are learnt using datasets containing data of one or both types. Dataset containing trajectories paired with $\langfeedback$, $\ddemo = \{(\tau_1,\langfeedback_1,\humano_1,\state_1),...(\tau_n,\langfeedback_n,\humano_n,\state_n)\}$,  and costmaps paired with $\langfeedback$,
$\dcmap = \{(\cmap_1,\langfeedback_1,\humano_1,\state_1),...(\cmap_k,\langfeedback_k,\humano_k,\state_k)\}$. Using $\mathcal{D}_{demos}$ and $\mathcal{D}_{cmap}$ we generate a unified final dataset.

\subsubsection{Generating Ground-truth $\cmap$ and $\mask$}
The $\cmap$ associated with trajectories $\tau$ in $\ddemo$ 
is generated by mapping every $\state$ on the trajectory to its distance to the goal measured along the trajectory and every $\state$ outside the trajectory is mapped to a fixed high cost. 
This kind of a mapping is representative of the fact that moving along the trajectory is definitively indicative of a decrease in cost for the specified $\langfeedback$. For tasks where cost maps are specified, for instance \emph{stay away from X}, the cost maps available are used directly. 
The process of generating ground truth masks for training is fairly straight forward. For datapoints in $\ddemo$, the binary mask is 1 for states along $\tau$ and is zero everywhere else. For datapoints in $\dcmap$ the mask is $I$. The final dataset is $\mathcal{D} = \{(\cmap_1,\mask_1,\langfeedback_1,\humano_1,\state_1),...\}$ where $c$ and $m$ correspond to the cost map and binary mask corresponding to the datapoint.

\subsubsection{Objective}
$\cmap$ and $\mask$ are learnt on the dataset $\mathcal{D}$ via maximum likelihood estimation. We initialize the parameters for models that learn $\cmap$ and $\mask$ with parameters $\theta$ and $\eta$. The models for $\cmap$ and $\mask$ condition on $\langfeedback$, $\state$, and $\humano$. The probability of generating the correct $\cmap$ and $\mask$ can be decomposed as follows.
\begin{align}
    p(\cmap \mid \langfeedback_i,\state_i,\humano_i) &= \prod_{s} p \Big(\cmap(s) \mid \langfeedback_i,\state_i,\humano_i \Big)    \\
    p(\mask \mid \langfeedback_i,\state_i,\humano_i) &= \prod_{s} p \Big( \mask(s) \mid \langfeedback_i,\state_i,\humano_i \Big)
\end{align}

To update model parameters the optimization objective is,
\begin{align}
    \theta &= \underset{\theta}{\argmax} \sum_{i=1}^{n+k} \sum_{\state \in \statspace} \log  p_{\theta} \Big( \cmap_{i}(\state) \mid ((\langfeedback_i,\state_i,\humano_i),\state) \Big)\\
    \eta &= \underset{\eta}{\argmax} \sum_{i=1}^{n+k} \sum_{\state \in \statspace} \log p_{\eta} \Big(\mask_{i}(\state)\mid ((\langfeedback_i,\state_i,\humano_i),\state) \Big)
\end{align}

For datapoints from $\ddemo$ we only penalize the cost prediction model for $\state$ on the trajectory whereas for datapoints from $\dcmap$ we penalize the model for the entire costmap. This is because, for demonstrations we are only confident about costs along the trajectory. This partial supervision used while training allows the model to extrapolate and make guesses of the costs everywhere else in the map and as a result the cost maps generated are smoother. At inference, to obtain the $\cmap$ and $\mask$ corresponding to a new $\langfeedback$, $\humano$ and $\state$,
\begin{align}
    \cmap &= \underset{\cmap}{\argmax} ~ p_{\theta}(\cmap \mid \langfeedback_i,\state_i,\humano_i) \\
    \mask &= \underset{\mask}{\argmax} ~ p_{\eta}(\mask \mid \langfeedback_i,\state_i,\humano_i)
\end{align}

\subsubsection{Interfacing $\languagecost$ with the $\planner$}
We explore two corrections types that can be encoded by our~$\languagecost$; constraint addition and  goal specification. In the case of constraint addition, the constraints are added to the~$\planner$ permanently~(e.g., going faster, slower, staying away from an object). While optimizing, we keep track of the constraints in a set~$\mathcal{C}_\mathcal{LC}$ to enable accounting over multiple constraints. In the case of goal specification, there are two cases in which a goal may be specified, first, in the absence of any previous goal and second, as a way to correct the model by introducing intermediate goals. In the first case, there is no existing goal cost and $\languagecost$ becomes the only active goal cost alongside the other constraints.
In the second mode we deactivate the task cost~$\taskcost$ and wait until ~$\languagecost$ is within a threshold~$\epsilon$ before activating the original task cost~$\taskcost$ back again. This temporary activation mode is used where ~$\langfeedback$ specifies an intermediate goal directive. The $\mask$ along with the presence or absence of an existing $\taskcost$ is indicative of the mode of correction. A $\langfeedback$ with a $\mask == 1$ is always a constraint and that with a $\mask \neq 1$ is a goal directive. We interface our~$\languagecost$ with an optimization based controller~\cite{bhardwaj2021storm} to generate commands for the robot as shown in Algorithm~\ref{algo:control}.

\begin{algorithm}[h!]
\caption{$\languagecost$ interfaced with the $\planner$}\label{algo:plannerinterface}
\small
\SetAlgoLined
\textbf{Initialize} $\mathcal{C}_\mathcal{LC}=[]$, $\languagecost=0$,  $\starrobotcost = \taskcost + \costbasic + \mathcal{C}_\mathcal{LC}$\\
\While{task not done}{
$\state_t$ = get\_new\_state() // Get current robot state\;
\uIf{user feedback}{
  $\langfeedback$ = get\_user\_feedback();\\
  $\cmap$ = $p_{\theta}$ $(\langfeedback, \state_t, \humano_t)$ \\
  $\mask$ = $p_{\eta}$ $(\langfeedback, \state_t, \humano_t)$ // Run inference;
\uIf{$\mask == I$}
{$\mathcal{C}_\mathcal{LC}$.append($\cmap * \mask$) // Add to constraints list\;}
\Else{$\languagecost = \cmap * \mask$ \\
$\starrobotcost = \taskcost + \costbasic + \languagecost + \mathcal{C}_\mathcal{LC}$
}
\uIf{$\languagecost(\state_t)<\epsilon$}{
$\starrobotcost = \taskcost + \costbasic + \mathcal{C}_\mathcal{LC}$ // Original goal-cost is active\;}
}
$a_{t+1} = \planner(\starrobotcost, \state_t)$ // Optimize with planner\;
command\_robot($a_{t+1}$) // Send command to robot\;
$t$+=1
}
\label{algo:control}
\end{algorithm}

\begin{figure*}[hbt!]
    \centering
    \includegraphics[width=\textwidth]{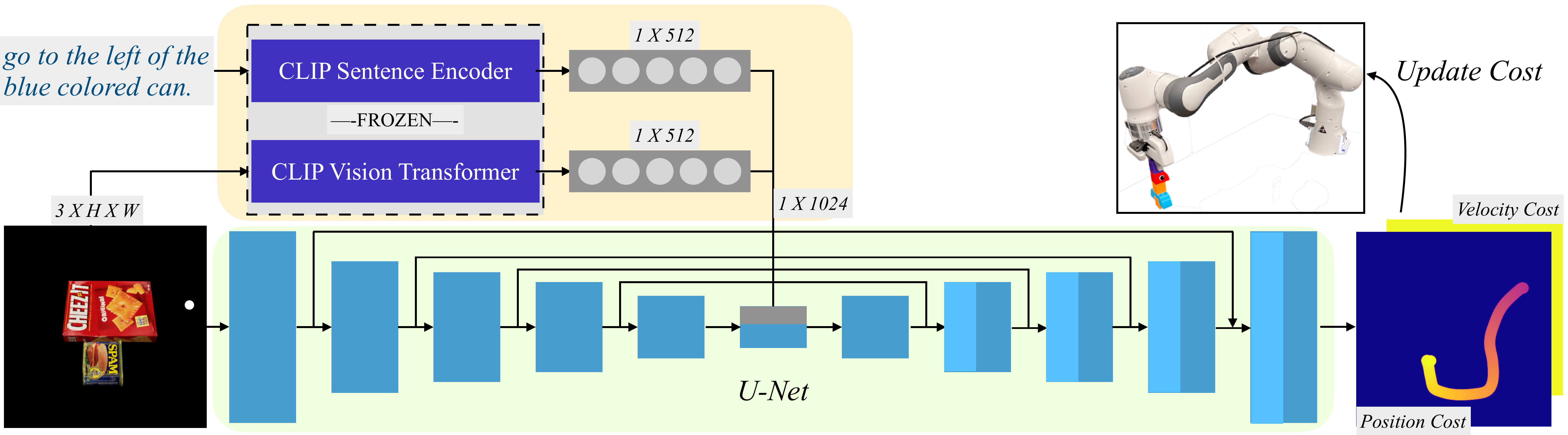}
    \caption{\textbf{Architecture:} The architecture of the language parametrized cost correction module consists of two streams
    .
    The CLIP stream takes as input the natural language feedback as well as an image of the environment  and the U-Net stream encodes the image of the environment. The output of this model is used to map to the cost associated with the language instruction. This could be learnt using specified costs corresponding to specific instructions or via estimating the cost from demonstrations. More details in section \cref{approach}}
    \label{fig:arch}
\end{figure*}

\subsection{Architecture}
$\cmap$ and $\mask$ are both individually implemented as a neural network with a two-stream architecture as seen in Fig \ref{fig:arch}. The CLIP stream consists of a pre-trained Contrastive Language-Image Pre-training (CLIP)~\cite{Radford2021LearningTV} model with the Vision Transformer(ViT)~\cite{Dosovitskiy2020ImageIW} visual module. It takes as input the language correction $\langfeedback$, robot state $\state$ and the RGB representation $\humano$. The state of the robot is encoded using the location of it's end effector on a spatial map of the RGB $\humano$. We use the $512$ dimensional visual embedding output from ViT along with the $512$ dimension language embedding output from the language-transformer. The image is encoded using a U-Net architecture with skip connections \cite{Ronneberger2015UnetCN}. It encodes the RGB image $\humano$ and robot state $\state$ of the robot and generates the output frames corresponding to the position cost map and velocity cost map each parametrized as 2D map in $\mathcal{R}^{\lvert \statspace \rvert}$\footnote{$\lvert \statspace \rvert$ denotes the dimension of the $\statspace$ }. The visual and language embedding from CLIP are concatenated with the embedding of the encoder before passing the embedding through the deconvolution layers to generate the cost map. 
The weights of the CLIP language transformer and ViT are frozen and the training optimizes the weights of the U-Net only. The model outputs two cost maps: 1) a position map and 2) a velocity map.\footnote{Our framework can also be extended to output additional maps for new quantities such as force and torque by adding additional frames to the output.}

\section{Experimental Protocol}
\label{sec:dataset}

\begin{figure*}[ht]
    \centering
    \includegraphics[width=\textwidth]{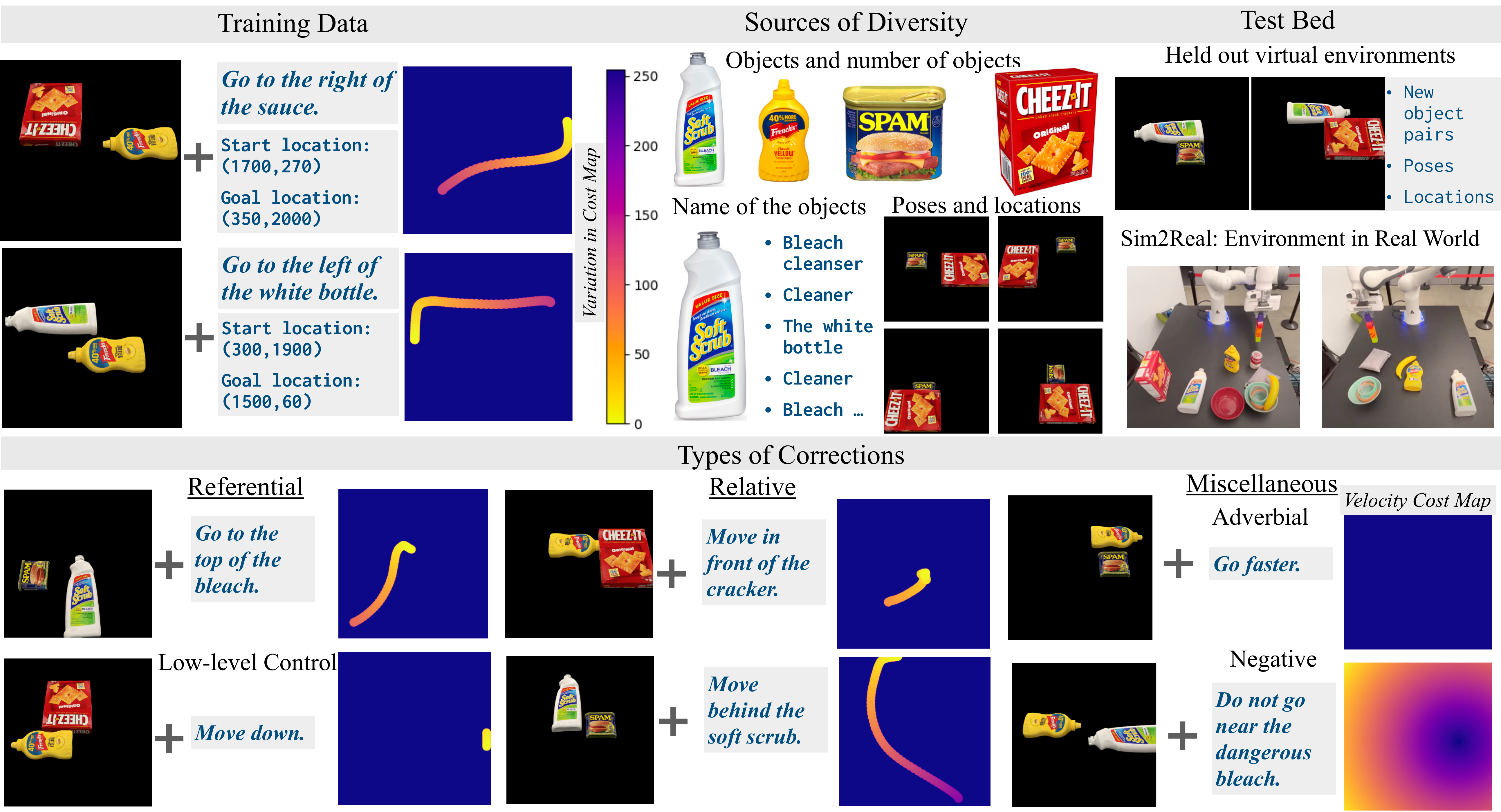}
    \caption{\textbf{Dataset:} Clockwise starting from top left a) The training data consists of data of the form (input:(instructions, environments,robot state), output:(cost)). The cost can be either specified by the user directly or come from demonstrations. When learnt from the demonstration, the trajectory is modified by linearly decreasing the cost from the start location to the end location along the trajectory. b) The sources of diversity in the dataset comes from the types of objects, randomized positioning of the objects in the environment and from instructions templated to refer to the same object in a variety of different ways as shown c) There are two testing setups. Unseen environments with new object combinations, object locations in simulation and a real environment setup with a 7DoF robot arm. d) The vocabulary of corrections include referential expressions, relative clauses and miscellaneous interesting correction types such as avoid the object or costs that modify non-positional attributes of the robot's behavior such a velocity.}
    \label{fig:dataset}
\end{figure*}

For $\ddemo$, we generate a dataset containing 100 environments. Each environment contained two objects from a set of four YCB objects ~\cite{calli2015ycb}: a Cheeze-it box, a bleach bottle, a can of spam and a bottle of mustard. The position of each object is sampled uniformly within the bounds of the environment. We sample object orientation from one of four equally-spaced options. We render each environment with the NVIDIA Scene Imaging Interface (NViSII)~\cite{morrical2021nvisii}. Images are top-down and are of size $(2048,2048)$ pixels. For every environment, we uniformly sample 10 different collision-free start positions. We choose goal positions to be 20 pixels offset from the mid-point of object edges where the offset is away from the object, and generate corresponding templated language instructions. This templated language is sampled from diverse referring expressions and object descriptions, as shown in Fig.~\ref{fig:dataset}. 

We use STORM~\cite{bhardwaj2021storm} as a planner which minimizes a Euclidean distance cost to generate trajectories from start to goal position. However, as this is not a global optimizer, it can get caught in local minima. Trajectories that successfully connect these positions are categorized as successful; failed trajectories are stored separately as a~\emph{hard} set for evaluation. In our setting, the planner failed 6.4\% of the time. More details on the planner is described in appendix~\ref{app:mpc}. We then divided the environments with successful trajectories into training, validation, and test sets, so that a specific object configuration will only appear in one split. 

For $\dcmap$, the cost maps are generated in the following way. For velocity speed-up, slow down, and when no constraints are given over velocity the velocity costmap outputs $0$, $1$, or $2$ respectively, corresponding to all $\state \in \statspace$. For instructions of avoiding objects the cost map generated is $-\sqrt{(s-c)^2}$ where c is the location of the centre of the object to be stayed away from. All cost values are re-scaled between $[0,255]$.

\textbf{Evaluation and metrics.}
We consider a trial a success when the robot reaches within 20 pixels of the goal position. We evaluate our method on two platforms: 1) using the planner on the test set in simulation and 2) using the planner on a real Franka Panda robot in cluttered environments that also contains unseen objects to study generalization.

\section{Results}
We will first discuss the effect of the different components in our model, followed by the performance of our method in Section.~\ref{sec:main_results}. We will then discuss the effectiveness of different types of feedback in Sec.~\ref{sec:feedback_type} and our generalization experiments in Sec.~\ref{sec:generalization}. We also show failures in Sec.~\ref{sec:failures}.

\subsection{Goal Reaching with Language}
\label{sec:main_results}

First, we test how our method can be leveraged to reach positions given directly as (x,y) to the planner, when the robot is stuck in a local minimum.
We evaluate on the test dataset where the planner was~100\% successful and also on the \emph{hard} set where the planner had 0\% success. Visualizations of some of these hard environments can be found in \cref{app:hard_envs}. 
With just a single language correction~(\emph{Single-correction}) we can improve success rate from 0\% to 81\% in the \emph{hard}-set of problems, which also brings our success rate to 98\% on the full test set. With another language correction~(\emph{Two-corrections}, we get our success rate to 93\% and 99\% in the \emph{hard} and test sets respectively. In this way, we see that minimal human input can bring the overall reliability of the system to an impressive level. 

We additionally tested our network's ability to understand language by starting the robot at the initial position and specifying the goal solely via the language string~(\emph{Goal-as-Language}). In this setting, we do not give the planner access to the desired (x,y) position and as such the success rate drops to 65\% on the full set. However, we see that in problems where the original planner failed to reach with access to (x,y)~(\emph{hard} set), we see that our network is able to succeed in 29\% of the set without requiring access to (x,y). Through the results in Table.~\ref{tab:correction}, we can see that the $\languagecost$ model pulls the robot in most situations with one or two corrections. A full list of environments, corrections provided, cost maps produced and trajectories can be found on the website~\href{https://sites.google.com/view/language-costs}{sites.google.com/view/language-costs}.

\begin{table}[bth!]
\centering
{%
\begin{tabular}{lrr}
\toprule
\multirowcell{2}{Model}& \multicolumn{2}{c}{Success Rate}  \\ 
\cmidrule{2-3}
& Hard & Solvable + Hard \\
\hline
Planner & 0\% & 94\% \\
Single-correction             & 81\% & 98\% \\
Two-corrections                & 93\% & 99\% \\
\midrule
Goal-as-Language & 29\% & 65\%\\
\bottomrule
\end{tabular}}
\caption{Performance on Goal Reaching
}
\label{tab:correction}
\end{table}

\subsection{Ablation Experiments}
\label{sec:ablations}
To evaluate the effect of different components of our model, we run our method in simulation with our \emph{solvable} test set of situations (independent and identically distributed with the training data).
Again, we specify the goal only as a language instruction but do not give the planner the $(x, y)$ position of the goal.
This enables us to quantify the performance of the mapping from language string to planner success without any bias from a goal cost.

We first disable the language module so that our network doesn't take any language input~(\emph{No-Language}). This is an under-specified system as the network does not know what the user feedback is and hence the success rate is only~4\%  as seen in table.~\ref{tab:ablations}. This ablation shows that our dataset is not biased and it indeed requires language input for success.

We then removed the vision input to the network~(\emph{No-Vision}), so both CLIP ViT and the U-Net encoder do not get the environment image or the location of the robot. This is done to test if simply given a language instruction, how well does exhibiting an average behavior do and the success rate is only at ~33\%. This success rate includes only very basic commands, like ``go left'', ``go down'', ``speed up'', et cetera; without vision, the system cannot accomplish any task that refers to an object.
When we remove our U-Net encoder~(\emph{No-U-Net-Encoder}) and only use input to CLIP, the network does not do any better than~(\emph{No-Vision}), implying that the 512 vision embedding from CLIP is not sufficient to encode our task specific environments.

Finally, we remove only the trajectory mask~$\mask$, and only use~$\cmap$ as cost in~\emph{No-Mask}. We see that this brings up the success rate to~58\% but adding the trajectory mask get our method to 69\% on the test dataset. %

\begin{table}[h!]
\centering
{%
\begin{tabular}{l r r}
\toprule
Category            &  Success on Solvable & Object Reference\\
\hline
No-Language &  4\%  & 0\% \\
No-Vision  &  33\% & 0\%\\
No-U-Net-Encoder & 33\% & 0\% \\
Ours(No-Mask)      &      58\% & 37\%\\
\textbf{Cost and Mask (Ours)}         &    \textbf{69\%} & \textbf{52\%}\\
\bottomrule
\end{tabular}}
\caption{Effect of various model ablations on performance, when given a goal language instruction and no Euclidean distance cost. We see that our proposed method which predicts both a cost and valid-area mask outperforms the alternatives. Figure~\ref{fig:convergence} in the appendix shows the rate convergence to the goal. It can be seen that the difference between performance of the No-mask and our model is on the medium and long trajectories.}
\label{tab:ablations}
\end{table}

\begin{figure*}[th]
    \centering
    \includegraphics[width=\textwidth]{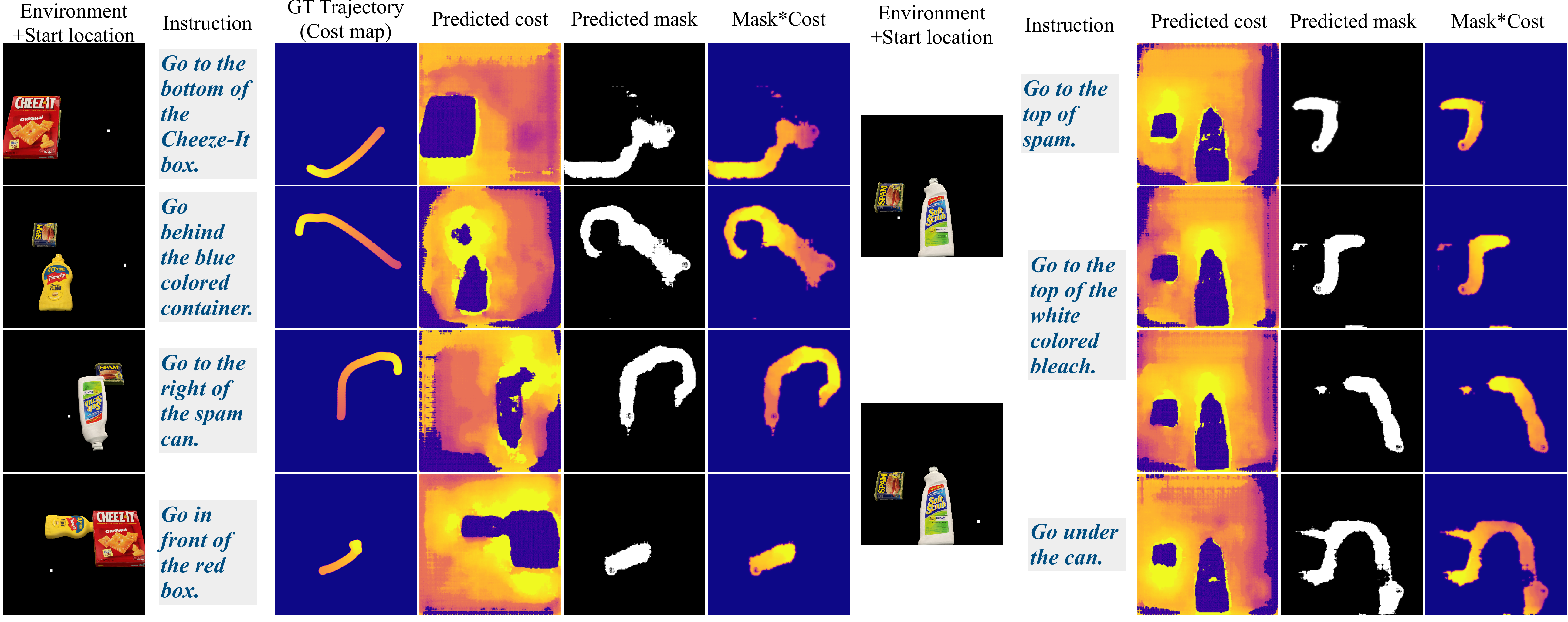}
    \caption{%
    The predicted cost map and generated trajectory together create a masked path for the robot to follow. The costs could also be used in isolation from the mask, though this results in worse performance. (Right) The model is sensitive to the environment, the start location of the robot and the instruction. Here, we show the costmap and masked costmap change as we hold other variables constant.}
    \label{fig:results-examples}
\end{figure*}

\subsection{Performance by Instruction Type} %
\label{sec:feedback_type}
In this section, we analyze our model's performance when given specific types of instruction. We trained on five spatial object dependent tasks-[Above, Below, Left of, Right of, Stay Away], two spatial robot object dependent tasks-[Behind, In front of], 4 directional spatial robot dependent tasks and 2 velocity tasks~(fast, slow). Table~\ref{tab:goal_results} shows a performance breakdown on each of these tasks. 
We evaluate on the solvable set of successful examples and split the trajectories by the total length of the ground-truth trajectory to analyze whether performance is dependant on the trajectory length. Specifically, paths of planning time steps $< 40$ were short, $40 - 60$ medium, and $>60$ were long. There were 304 short examples, 131 medium examples, and 108 long examples. When running these evaluations, we do not give access of $(x,y)$ goal positions to the planner and only give a language string specifying where the robot needs to go. The overall success rate was $69\%$; we see that specifically very local corrections such as ``go faster'' were always successful.
Fig.~\ref{fig:results-examples} shows the interaction between environment, cost prediction, and mask prediction for various goal instructions.

\begin{table}[h]
\centering
{%
\begin{tabular}{lrrrr}
\toprule[0.1em]
\multirowcell{2}{Type of\\Feedback}& \multicolumn{4}{c}{Success Rate on Solvable Set}  \\ 
\cmidrule{2-5}
& All & Short & Medium & Long \\ \toprule[0.1em]
Above               &                   77\%   & 86\%  & 83\%     & 40\%             \\
Below               &                      66\% & 70\%     & 77\% & 46\%            \\ 
Left of  &          45\%      & 82\% & 33\% & 12\%                        \\
Right of   &            49\% & 92\% & 47\% & 17\%            \\\midrule[0.05em]
Behind          &      20\%      & 17\% & 26\% & 18\%     \\
In front of   &   76\%      & 83\% & 50\% &  -                               \\\midrule[0.05em]
Positional($\updownarrow \leftrightarrow$)                                            &  100\%  & 100\% & - & -                                  \\
Velocity                                              &   100\%     & - & - & -                                   \\ \midrule[0.05em]
Stay Away &   95\%     & - & - & -                                   \\ 
\bottomrule[0.1em]
\end{tabular}}
\caption{How effective is the language cost prediction module at navigating to various types of goals? We look at problems of varying levels of difficulty. The types are partitioned by horizontal lines based on categories described in Sec~\ref{sec:feedback_type}}
\label{tab:goal_results}
\end{table}
\subsection{Generalization Experiments}
\label{sec:generalization}
The CLIP embeddings used in our model provide a strong basis for language generalization, as seen in previous work~\cite{thomason2022language,shridhar2022cliport}. We performed additional set of experiments to show how our models preserve this generalization ability, making them more broadly useful despite the small amount of training data on only four objects. We evaluate on the real robot for these results, where the environment also contains unseen objects in clutter. The results of these generalization experiments are in Table~\ref{tab:ood}, which shows how our approach can scale to a wide range of problems.

\begin{figure}[ht]
    \centering
    \includegraphics[width=0.47\textwidth]{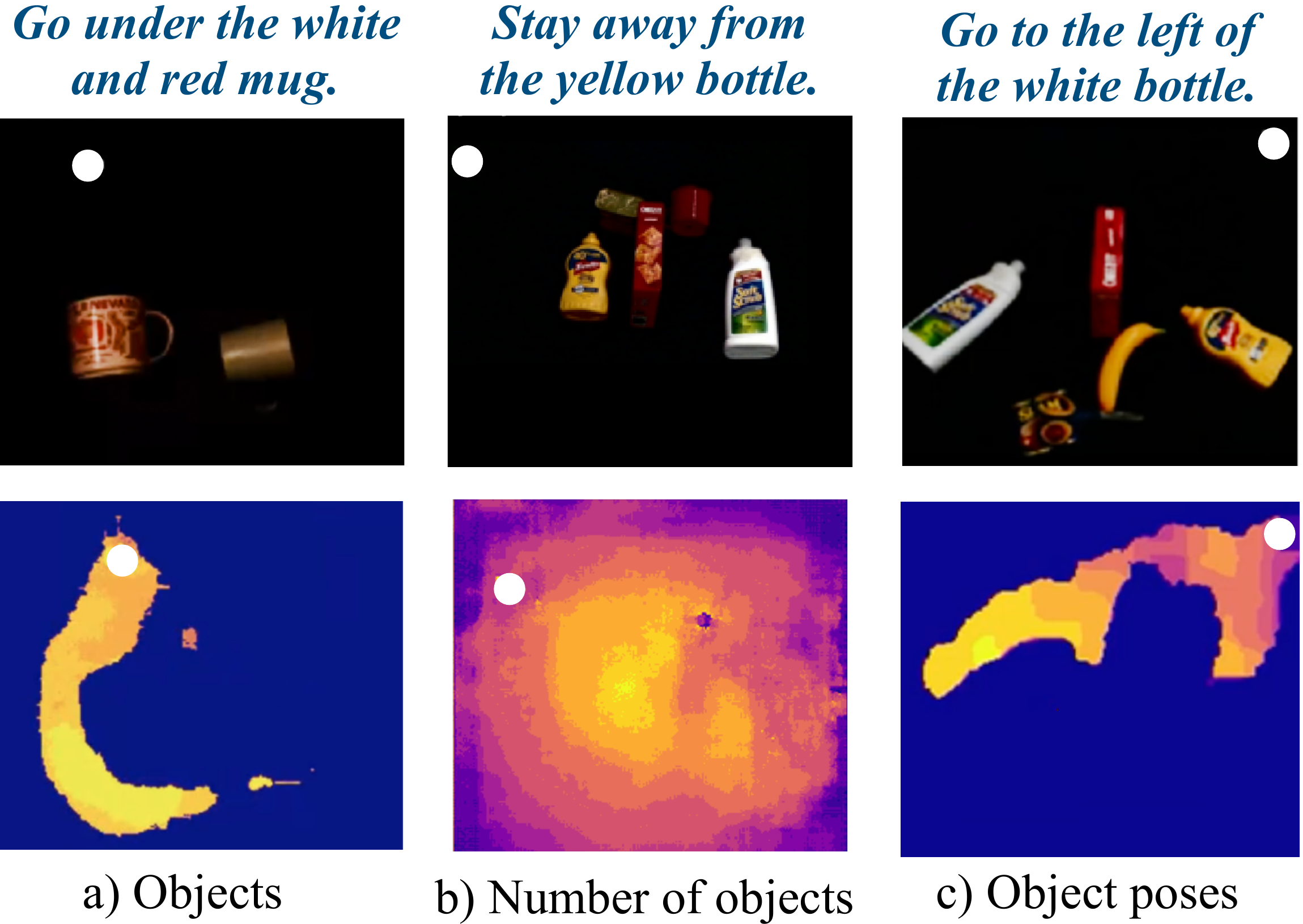}
    \caption{\textbf{Visual Generalization:} Visually the model generalized to unseen objects in the scene, to a different number of objects in the scene (as compared to 2 at training time) and to new poses of objects.}
    \label{fig:vis_general}
\end{figure}

To study generalization to unseen language instructions, we referred to the objects with non-templated phrases and synonymous object names that were not part of the training vocabulary, in 20 different setups and found that our method was able to successfully complete the task in 17 of them. Fig.~\ref{fig:languagediversity} shows some examples of diverse natural-language sentences used to control the robot. We also tested our method with language instructions referring to 10 unseen objects and our method worked on~\emph{cups(red, orange), plate, fork, ketchup bottle} and failed on~\emph{screwdriver, candle, book, banana, pepper can, pen}. When any of the unseen objects were in the background~(clutter) and language instructions were in reference to seen objects, our method succeeded always even when the objects were placed in orientations that were not seen during training as shown in Fig.~\ref{fig:vis_general}. Our training data only contained scenes with two of four objects in poses chosen from a fixed set of 4 orientations while the evaluations we did on the real robot contained many objects and seen objects in different orientations.
\begin{table}[bth!]
\centering
{%
\begin{tabular}{lr}
\toprule
Generalization type            & Task Success \\
\hline
New language instructions &  17/20 \\
Reference to unseen objects &  4/10 \\
New objects in background; clutter      &      15/15 \\
Unseen poses for known objects            &    14/15 \\
\bottomrule

\end{tabular}}
\caption{
Performance on generalization experiments. Our approach can generalize to new objects, language instructions, and to clutter that was not present in the training dataset.
}
\label{tab:ood}
\vskip -0.5cm
\end{table}

\begin{figure*}[ht]
    \centering
    \includegraphics[width=\textwidth]{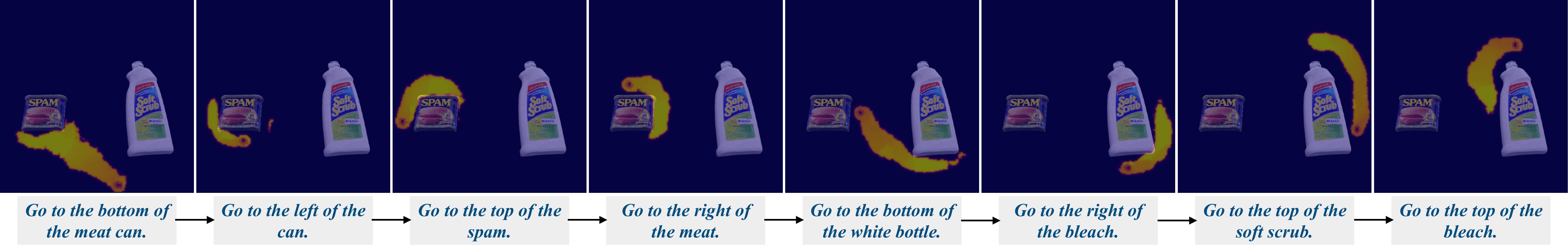}
    \caption{Compositionality in time. Our approach allows the user to specify multiple different cost functions at different points in time, letting them guide the robot around an intended trajectory with language.}
    \label{fig:timecomp}
\end{figure*}

\begin{figure*}[ht]
    \centering
    \includegraphics[width=\textwidth]{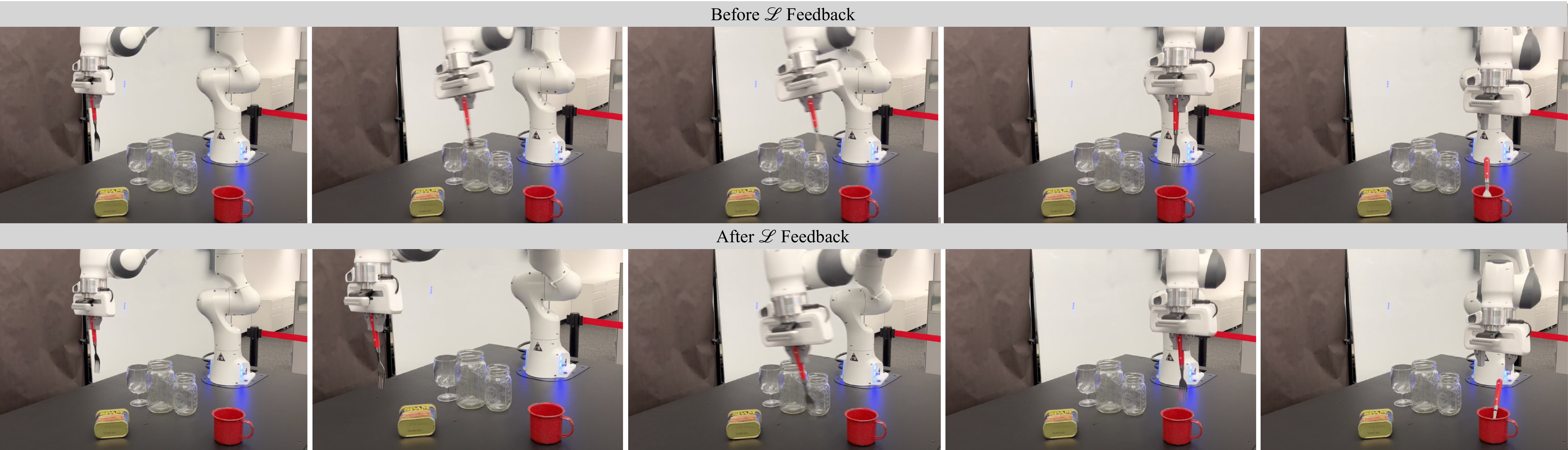}
    \caption{An example using our framework to tell a robot to avoid a set of fragile, glass objects while performing a pick and place task. Before, the robot moves dangerously close to the glass containers; after ``go through the bottom of spam'', it avoids them and maintains a safe distance.}
    \label{fig:preference}
\end{figure*}

\begin{figure}[ht]
    \centering
    \includegraphics[width=0.48\textwidth]{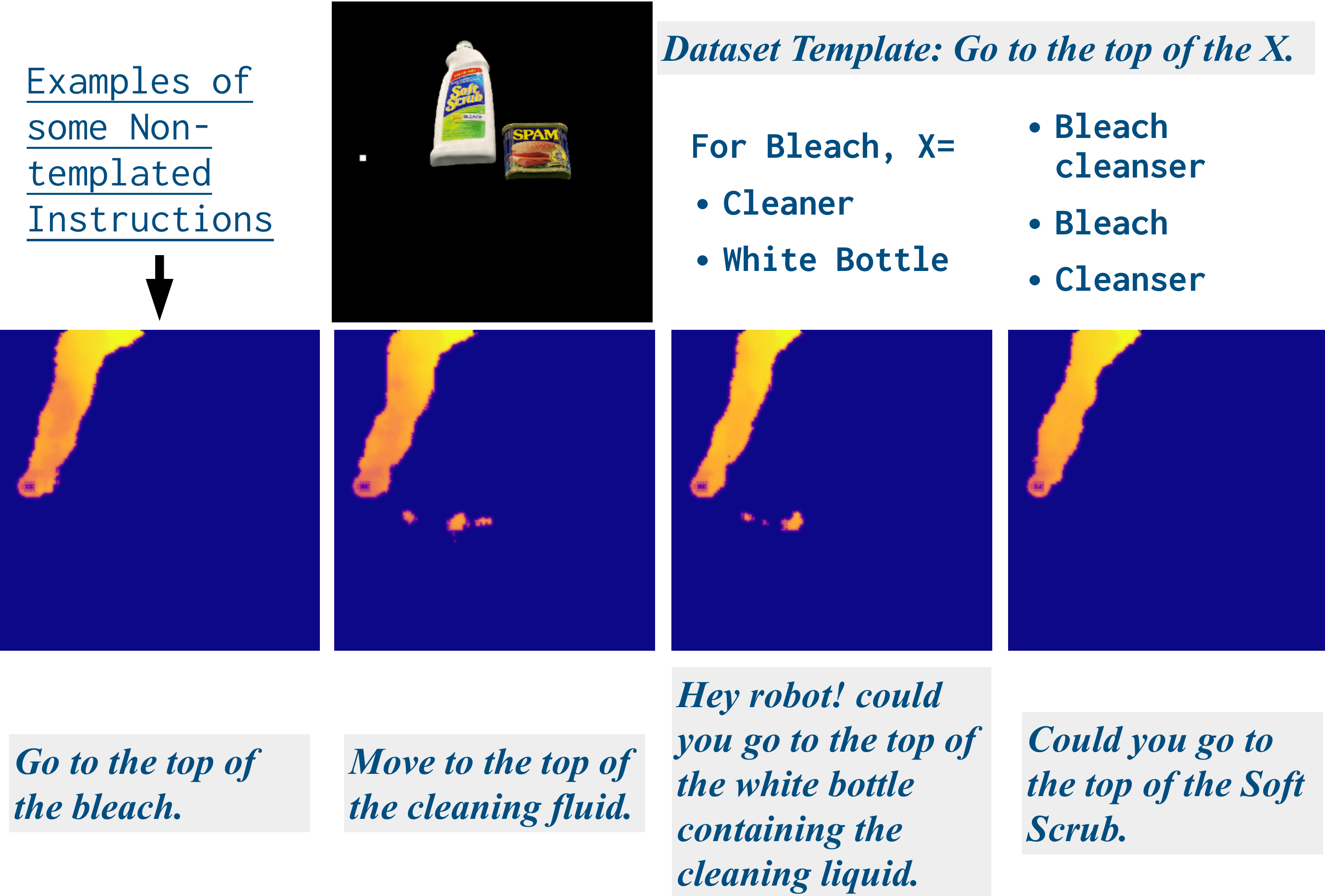}
    \caption{The language correction module is trained with templated language instructions.
    For example, to teach the robot to go over an object, we use the template ``Go to the top of the X,'' where X is the name of an object that can be referred to in several ways.
    However, in spite of this, our model can generalize to a wide variety of different types of instructions, as shown above.
    }
    \label{fig:languagediversity}
\end{figure}

\subsection{Failures}
\label{sec:failures}
\begin{figure}[ht]
    \centering
    \includegraphics[width=0.48\textwidth]{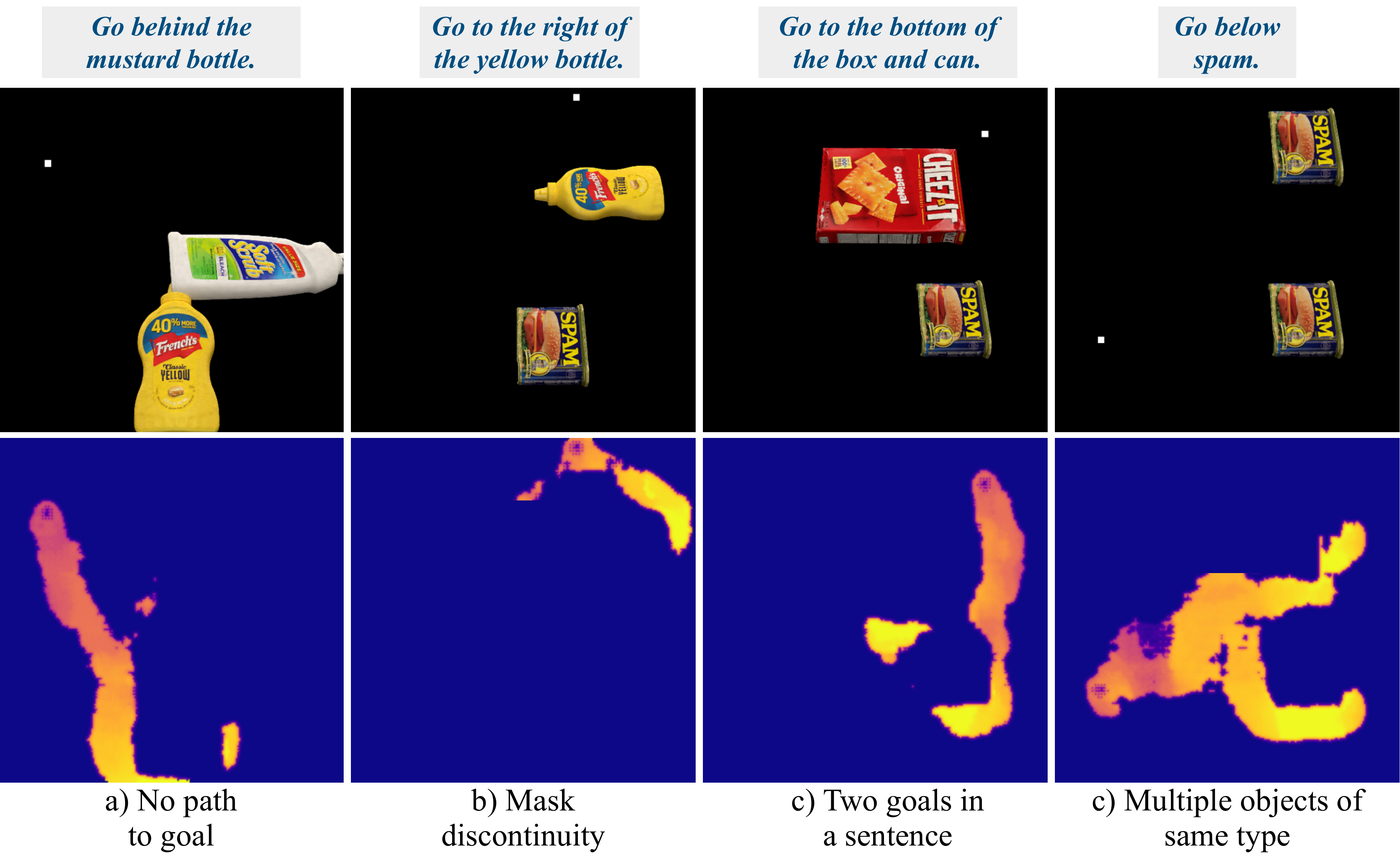}
    \caption{Some failures of our approach is shown here. Our method can produce masks that go outside the bounds of the image~(left) or discontinuous masks in some instances. Our method also cannot distinguish between two objects of the same type as shown in the right.}
    \label{fig:failures}
\end{figure}

Most failures of the correction policy are either due to a discontinuity in the trajectory mask generated or due to some few-off pixels in the cost map along the trajectory. Examples of these can be seen in Fig~\ref{fig:failures}. The figure also describes other curious scenarios. First, in the absence of an observable path $\languagecost$ tries to find a path from the edges of the frame. In the case of environments with two instances of the same object the model generates two distinct paths to both the objects. This is also true in cases when there are two objects and there is ambiguity in the instruction as seen in Fig~\ref{fig:failures}. 

\subsection{Discussion}

In addition to generalization, our approach has the advantages of compositionality over other means of providing feedback to improve robot behavior.
We can specify multiple constraints at execution time and combine them in the $\languagecost$ term. These can either be provided at once or at different times through the trial. We provide demonstrations showing a robot’s behavior at combining velocity cost with goal reaching costs and with stay away cost while the robot is trying to reach a goal on our website~\href{https://sites.google.com/view/language-costs}{sites.google.com/view/language-costs}

Issuing commands at different times is also a powerful and intuitive way to control the robot. For example, in Fig.~\ref{fig:timecomp} we see the robot being asked to go to different locations around objects sequentially.  Chaining corrections makes possible specifying procedures, and correcting trajectories that require more than a single correction to be corrected. It might also provide a way to generate data to learn more complex behaviors.

\begin{figure}[ht]
    \centering
    \includegraphics[width=0.48\textwidth]{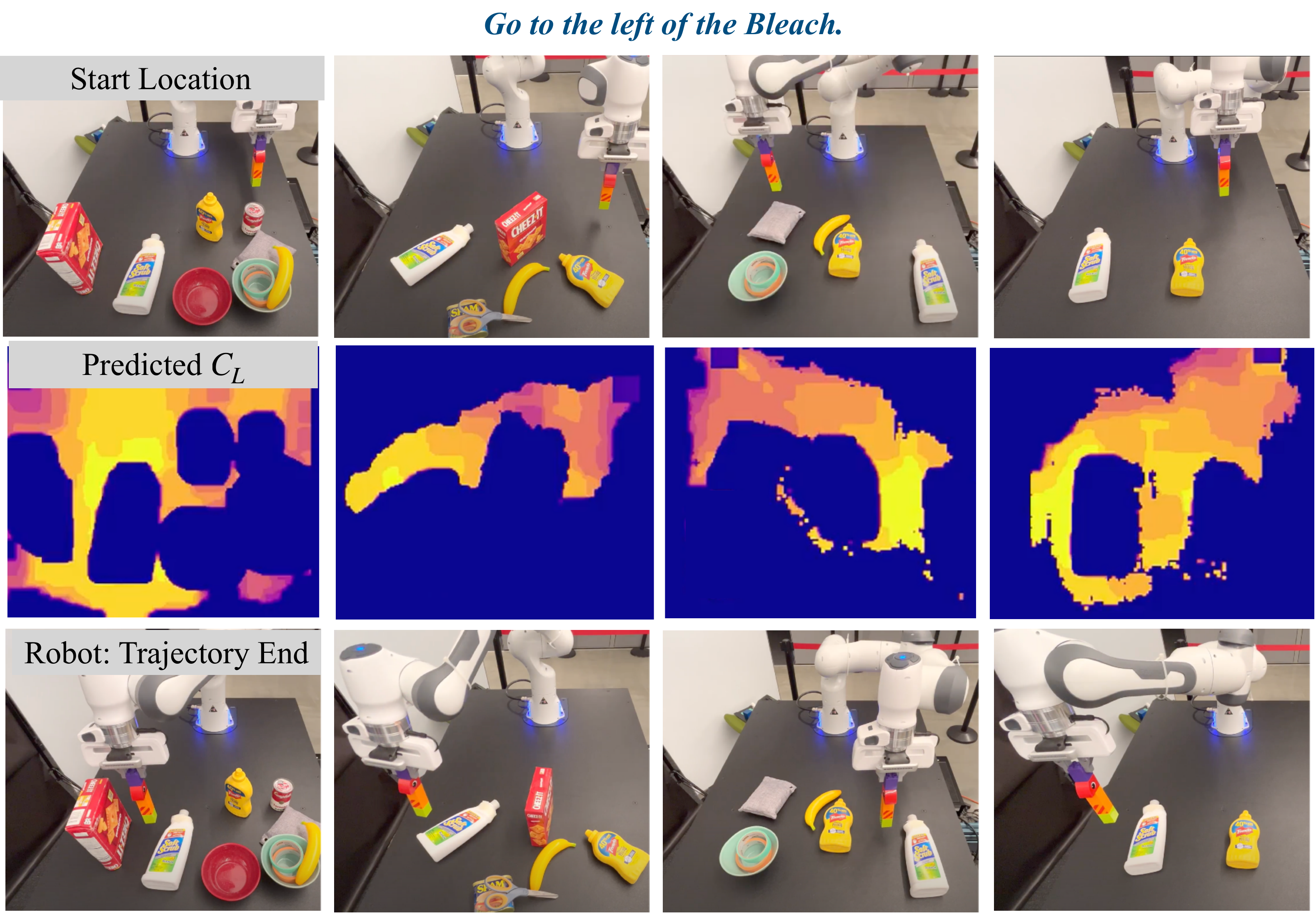}
    \caption{Providing a single language command in multiple environments. One significant advantage of using language for correcting robot behaviors is how it allows us to issue the same correction in many environments at once.}
    \label{fig:bleach-left}
\end{figure}

One useful feature of our approach is the ability to correct behaviors across multiple environments at once.
Take the example in Fig.~\ref{fig:bleach-left}: ``go to the left of the bleach.'' This correction can be applied in every environment, even if the robot is moving to different goals, with no additional effort on the part of the user.
To make such a correction with other means such as a joystick or kinesthetic feedback would require considerable time and effort. %

Importantly, corrections do not always need to be provided \emph{only} in the case of a planner failure, but can be used by the user to modify the plan based on their \emph{preference}. In Figure~\ref{fig:preference} we see a human providing a correction to steer the robot away from fragile objects. This correction is applied to the task of placing an object in the mug.

\section{Related Work}
\label{sec:related_work}

This work builds upon multiple threads of related work.

\subsubsection{Natural Language for Robot Behavior Correction}
Correcting robot behavior using language has been studied for robots that ask for help~\cite{Tellex2014AskingFH}, understand language corrections~\cite{Broad2016TowardsRT, Co2018GuidingPW}, and use language to disambiguate joystick corrections~\cite{Karamcheti2022LilaLI}.
~\citet{Broad2016TowardsRT} use a grounding model based on a Distributed Correspondence Graph~\cite{Howard2014NaturalLP} to ground corrections, which limits grounding language to a hand-engineered set of optimization constraints.
In contrast, we directly learn to predict cost maps using a neural network, side-stepping the potentially laborious constraint design process.
~\citet{Co2018GuidingPW} learn a policy that accepts corrections in addition to an instruction, however it requires training with corrections at training time which makes it  sample inefficient as number of tasks scale.
Further, it does not permit decoupling the notions of a goal and a trajectory to said goal when issuing a correction.
~\citet{Karamcheti2022LilaLI} use language to disambiguate underspecified joystick corrections by learning a mapping to robot joint space. This suffers from the same limitations as language-free joystick~\cite{Spencer-RSS-20,rakita2018autonomous} or kinesthetic~\cite{bajcsy-hri2018} corrections, requiring undivided user attention. Furthermore, their language grounding model is based on a nearest-neighbour lookup, which does not generalize to new environments and tasks.

\subsubsection{Language for Robot Task Specification}
Natural language has been extensively studied as a means for task specification or instruction following in robotics~\cite{Tellex2011UnderstandingNL, Matuszek2012LearningTP, Duvallet2013ImitationLF, Arumugam2017AccuratelyAE, Nyga2018GroundingRP, Williams2018LearningTP, Blukis2019LearningTM, Paxton2019ProspectionIP}.
Mapping language to symbolic plans~\cite{Tellex2011UnderstandingNL, Matuszek2012LearningTP, Misra2014ContextSG, Arumugam2017AccuratelyAE}, has enabled following instructions by invoking a set of pre-specified skills or motion primitives.
Recently, instruction following has been studied in robotics by mapping instructions and raw visual observations to actions using end-to-end representation learning and sim-to-real transfer~\cite{Blukis2019LearningTM, Anderson2020SimTR, shridhar2022cliport}.
All of these works, however, treat language instructions as goals that are fixed during execution.
In contrast, by framing language grounding as cost prediction, we enable the use of language to refine robot behavior over time, while still allowing instruction-following as a special use-case.

\subsubsection{Inferring costs from demonstrations}
Our correction model is trained to map observations and language to cost maps, on data consisting of demonstrations or ground-truth cost maps.
This is related to Inverse Reinforcement Learning (IRL)~\cite{Abbeel2004ApprenticeshipLV} that learns to recover reward functions from demonstrations, and has been successfully applied to infer objectives for manipulation motion planners~\cite{Mrinal2013LearningOF}.
In contrast, our cost model is conditioned on language and visual observations, which enables re-using the same model with diverse language corrections in a variety of tasks.

\subsubsection{Value alignment problem} 
Our method presents an interactive solution to the value alignment problem~\cite{Wallach2008MoralMT}, whereby the cost function provided to the robot is not reflective of the true task that the user has in mind. Our language corrections enable interactively updating the cost to better reflect the task.
This problem has been also addressed by learning to predict true rewards given observed rewards across environments and tasks~\cite{HadfieldMenell2017InverseRD}, and by learning from physical interactions with humans~\cite{Bajcsy2017LearningRO}.

\section{Conclusion}

In conclusion, we proposed a framework to integrate human provided feedback in natural language to update a robot's planning cost applied to situations when the planner fails. This is done by modelling cost associated with the language instruction  $\languagecost$ conditioned on the language feedback $\langfeedback$. The $\languagecost$ can be used in conjunction with the motion planner's existing costs. We train our model on data generated via simulation and evaluate the performance of the model in various out of distribution settings in the real world involving non-templated natural user commands, cluttered scenes, new poses and types of objects.

\bibliographystyle{plainnat}
\bibliography{references}

\begin{thebibliography}{44}
\providecommand{\natexlab}[1]{#1}
\providecommand{\url}[1]{\texttt{#1}}
\expandafter\ifx\csname urlstyle\endcsname\relax
  \providecommand{\doi}[1]{doi: #1}\else
  \providecommand{\doi}{doi: \begingroup \urlstyle{rm}\Url}\fi

\bibitem[Abbeel and Ng(2004)]{Abbeel2004ApprenticeshipLV}
Pieter Abbeel and Andrew~Y Ng.
\newblock Apprenticeship learning via inverse reinforcement learning.
\newblock In \emph{ICML}, 2004.

\bibitem[Anderson et~al.(2020)Anderson, Shrivastava, Truong, Majumdar, Parikh,
  Batra, and Lee]{Anderson2020SimTR}
Peter Anderson, Ayush Shrivastava, Joanne Truong, Arjun Majumdar, Devi Parikh,
  Dhruv Batra, and Stefan Lee.
\newblock Sim-to-real transfer for vision-and-language navigation.
\newblock In \emph{CoRL}, 2020.

\bibitem[Andreas and Klein(2015)]{Andreas2015AlignmentBC}
Jacob Andreas and Dan Klein.
\newblock Alignment-based compositional semantics for instruction following.
\newblock In \emph{EMNLP}, 2015.

\bibitem[Arumugam et~al.(2017)Arumugam, Karamcheti, Gopalan, Wong, and
  Tellex]{Arumugam2017AccuratelyAE}
Dilip Arumugam, Siddharth Karamcheti, Nakul Gopalan, Lawson~LS Wong, and
  Stefanie Tellex.
\newblock Accurately and efficiently interpreting human-robot instructions of
  varying granularities.
\newblock In \emph{RSS}, 2017.

\bibitem[Bajcsy et~al.(2017)Bajcsy, Losey, O'Malley, and
  Dragan]{Bajcsy2017LearningRO}
Andrea Bajcsy, Dylan~P. Losey, Marcia~Kilchenman O'Malley, and Anca~D. Dragan.
\newblock Learning robot objectives from physical human interaction.
\newblock In \emph{CoRL}, 2017.

\bibitem[Bajcsy et~al.(2018)Bajcsy, Losey, O’Malley, and
  Dragan]{bajcsy-hri2018}
Andrea Bajcsy, Dylan~P. Losey, Marcia~K. O’Malley, and Anca~D. Dragan.
\newblock Learning from physical human corrections, one feature at a time.
\newblock In \emph{ACM/IEEE International Conference on Human-Robot Interaction
  (HRI)}, pages 141--149, 2018.

\bibitem[Bhardwaj et~al.(2021)Bhardwaj, Sundaralingam, Mousavian, Ratliff, Fox,
  Ramos, and Boots]{bhardwaj2021storm}
Mohak Bhardwaj, Balakumar Sundaralingam, Arsalan Mousavian, Nathan~D Ratliff,
  Dieter Fox, Fabio Ramos, and Byron Boots.
\newblock Storm: An integrated framework for fast joint-space model-predictive
  control for reactive manipulation.
\newblock In \emph{5th Annual Conference on Robot Learning}, 2021.

\bibitem[Blukis et~al.(2019)Blukis, Terme, Niklasson, Knepper, and
  Artzi]{Blukis2019LearningTM}
Valts Blukis, Yannick Terme, Eyvind Niklasson, Ross~A. Knepper, and Yoav Artzi.
\newblock Learning to map natural language instructions to physical quadcopter
  control using simulated flight.
\newblock In \emph{CoRL}, 2019.

\bibitem[Bowyer et~al.(2013)Bowyer, Davies, and y~Baena]{bowyer2013active}
Stuart~A Bowyer, Brian~L Davies, and Ferdinando~Rodriguez y~Baena.
\newblock Active constraints/virtual fixtures: A survey.
\newblock \emph{IEEE Transactions on Robotics}, 30\penalty0 (1):\penalty0
  138--157, 2013.

\bibitem[Broad et~al.(2016)Broad, Arkin, Ratliff, Howard, Argall, and
  Graph]{Broad2016TowardsRT}
Alexander Broad, Jacob Arkin, Nathan Ratliff, Thomas Howard, Brenna Argall, and
  Distributed~Correspondence Graph.
\newblock Towards real-time natural language corrections for assistive robots.
\newblock In \emph{RSS Workshop on Model Learning for Human-Robot
  Communication}, 2016.

\bibitem[Calli et~al.(2015)Calli, Singh, Walsman, Srinivasa, Abbeel, and
  Dollar]{calli2015ycb}
Berk Calli, Arjun Singh, Aaron Walsman, Siddhartha Srinivasa, Pieter Abbeel,
  and Aaron~M Dollar.
\newblock The ycb object and model set: Towards common benchmarks for
  manipulation research.
\newblock In \emph{2015 international conference on advanced robotics (ICAR)},
  pages 510--517. IEEE, 2015.

\bibitem[Co-Reyes et~al.(2018)Co-Reyes, Gupta, Sanjeev, Altieri, Andreas,
  DeNero, Abbeel, and Levine]{Co2018GuidingPW}
John~D Co-Reyes, Abhishek Gupta, Suvansh Sanjeev, Nick Altieri, Jacob Andreas,
  John DeNero, Pieter Abbeel, and Sergey Levine.
\newblock Guiding policies with language via meta-learning.
\newblock In \emph{ICLR}, 2018.

\bibitem[Dosovitskiy et~al.(2020)Dosovitskiy, Beyer, Kolesnikov, Weissenborn,
  Zhai, Unterthiner, Dehghani, Minderer, Heigold, Gelly,
  et~al.]{Dosovitskiy2020ImageIW}
Alexey Dosovitskiy, Lucas Beyer, Alexander Kolesnikov, Dirk Weissenborn,
  Xiaohua Zhai, Thomas Unterthiner, Mostafa Dehghani, Matthias Minderer, Georg
  Heigold, Sylvain Gelly, et~al.
\newblock An image is worth 16x16 words: Transformers for image recognition at
  scale.
\newblock In \emph{International Conference on Learning Representations}, 2020.

\bibitem[Dragan and Srinivasa(2013)]{Dragan2013APF}
Anca~D. Dragan and Siddhartha~S. Srinivasa.
\newblock A policy-blending formalism for shared control.
\newblock \emph{The International Journal of Robotics Research}, 32:\penalty0
  790 -- 805, 2013.

\bibitem[Duvallet et~al.(2013)Duvallet, Kollar, and
  Stentz]{Duvallet2013ImitationLF}
Felix Duvallet, Thomas Kollar, and Anthony Stentz.
\newblock Imitation learning for natural language direction following through
  unknown environments.
\newblock \emph{2013 IEEE International Conference on Robotics and Automation},
  pages 1047--1053, 2013.

\bibitem[Hadfield-Menell et~al.(2017)Hadfield-Menell, Milli, Abbeel, Russell,
  and Dragan]{HadfieldMenell2017InverseRD}
Dylan Hadfield-Menell, Smitha Milli, P.~Abbeel, Stuart~J. Russell, and Anca~D.
  Dragan.
\newblock Inverse reward design.
\newblock In \emph{NIPS}, 2017.

\bibitem[Harnad(1990)]{Harnad1990SymbolGP}
Stevan Harnad.
\newblock The symbol grounding problem.
\newblock \emph{Physica D: Nonlinear Phenomena}, 1990.

\bibitem[Howard et~al.(2014)Howard, Tellex, and Roy]{Howard2014NaturalLP}
Thomas~M Howard, Stefanie Tellex, and Nicholas Roy.
\newblock A natural language planner interface for mobile manipulators.
\newblock In \emph{2014 IEEE International Conference on Robotics and
  Automation (ICRA)}, pages 6652--6659. IEEE, 2014.

\bibitem[Jain et~al.(2015)Jain, Sharma, Joachims, and
  Saxena]{Jain2015LearningPF}
Ashesh Jain, Shikhar Sharma, Thorsten Joachims, and Ashutosh Saxena.
\newblock Learning preferences for manipulation tasks from online coactive
  feedback.
\newblock \emph{The International Journal of Robotics Research}, 34:\penalty0
  1296 -- 1313, 2015.

\bibitem[Kalakrishnan et~al.(2013)Kalakrishnan, Pastor, Righetti, and
  Schaal]{Mrinal2013LearningOF}
Mrinal Kalakrishnan, Peter Pastor, Ludovic Righetti, and Stefan Schaal.
\newblock Learning objective functions for manipulation.
\newblock In \emph{ICRA}, 2013.

\bibitem[Karamcheti et~al.(2021)Karamcheti, Srivastava, Liang, and
  Sadigh]{Karamcheti2022LilaLI}
Siddharth Karamcheti, Megha Srivastava, Percy Liang, and Dorsa Sadigh.
\newblock Lila: Language-informed latent actions.
\newblock In \emph{CoRL}, 2021.

\bibitem[Matuszek et~al.(2012)Matuszek, Herbst, Zettlemoyer, and
  Fox]{Matuszek2012LearningTP}
Cynthia Matuszek, Evan Herbst, Luke Zettlemoyer, and Dieter Fox.
\newblock Learning to parse natural language commands to a robot control
  system.
\newblock In \emph{ISER}, 2012.

\bibitem[Misra et~al.(2014{\natexlab{a}})Misra, Sung, Lee, and
  Saxena]{misra_tellmedave_2014}
Dipendra Misra, Jaeyong Sung, Kevin Lee, and Ashutosh Saxena.
\newblock Tell me dave: Context-sensitive grounding of natural language to
  mobile manipulation instructions.
\newblock In \emph{Robotics: Science and Systems (RSS)}, 2014{\natexlab{a}}.

\bibitem[Misra et~al.(2014{\natexlab{b}})Misra, Sung, Lee, and
  Saxena]{Misra2014ContextSG}
Dipendra~K Misra, Jaeyong Sung, Kevin Lee, and Ashutosh Saxena.
\newblock Tell me dave: Context-sensitive grounding of natural language to
  mobile manipulation instructions.
\newblock In \emph{RSS}, 2014{\natexlab{b}}.

\bibitem[Mooney(2008)]{Mooney2008LearningTC}
Raymond~J Mooney.
\newblock Learning to connect language and perception.
\newblock In \emph{AAAI}, pages 1598--1601, 2008.

\bibitem[Morrical et~al.(2021)Morrical, Tremblay, Lin, Tyree, Birchfield,
  Pascucci, and Wald]{morrical2021nvisii}
Nathan Morrical, Jonathan Tremblay, Yunzhi Lin, Stephen Tyree, Stan Birchfield,
  Valerio Pascucci, and Ingo Wald.
\newblock Nvisii: A scriptable tool for photorealistic image generation, 2021.

\bibitem[Muxfeldt et~al.(2014)Muxfeldt, Kluth, and
  Kubus]{Muxfeldt2014KinestheticTI}
Arne Muxfeldt, Jan-Henrik Kluth, and Daniel Kubus.
\newblock Kinesthetic teaching in assembly operations - a user study.
\newblock In \emph{SIMPAR}, 2014.

\bibitem[Nyga et~al.(2018)Nyga, Roy, Paul, Park, Pomarlan, Beetz, and
  Roy]{Nyga2018GroundingRP}
Daniel Nyga, Subhro Roy, Rohan Paul, Daehyung Park, Mihai Pomarlan, Michael
  Beetz, and Nicholas Roy.
\newblock Grounding robot plans from natural language instructions with
  incomplete world knowledge.
\newblock In \emph{CoRL}, 2018.

\bibitem[Padmakumar et~al.(2018)Padmakumar, Stone, and
  Mooney]{Padmakumar2018LearningAP}
Aishwarya Padmakumar, Peter Stone, and Raymond~J. Mooney.
\newblock Learning a policy for opportunistic active learning.
\newblock In \emph{EMNLP}, 2018.

\bibitem[Paul et~al.(2016)Paul, Arkin, Roy, and M~Howard]{Paul2016EfficientOA}
Rohan Paul, Jacob Arkin, Nicholas Roy, and Thomas M~Howard.
\newblock Efficient grounding of abstract spatial concepts for natural language
  interaction with robot manipulators.
\newblock 2016.

\bibitem[Paxton et~al.(2019)Paxton, Bisk, Thomason, Byravan, and
  Fox]{Paxton2019ProspectionIP}
Chris Paxton, Yonatan Bisk, Jesse Thomason, Arunkumar Byravan, and Dieter Fox.
\newblock Prospection: Interpretable plans from language by predicting the
  future.
\newblock In \emph{ICRA}, 2019.

\bibitem[Radford et~al.(2021)Radford, Kim, Hallacy, Ramesh, Goh, Agarwal,
  Sastry, Askell, Mishkin, Clark, Krueger, and
  Sutskever]{Radford2021LearningTV}
Alec Radford, Jong~Wook Kim, Chris Hallacy, Aditya Ramesh, Gabriel Goh,
  Sandhini Agarwal, Girish Sastry, Amanda Askell, Pamela Mishkin, Jack Clark,
  Gretchen Krueger, and Ilya Sutskever.
\newblock Learning transferable visual models from natural language
  supervision.
\newblock In \emph{ICML}, 2021.

\bibitem[Rakita et~al.(2018)Rakita, Mutlu, and Gleicher]{rakita2018autonomous}
Daniel Rakita, Bilge Mutlu, and Michael Gleicher.
\newblock An autonomous dynamic camera method for effective remote
  teleoperation.
\newblock In \emph{Proceedings of the 2018 ACM/IEEE International Conference on
  Human-Robot Interaction}, pages 325--333, 2018.

\bibitem[Ronneberger et~al.(2015)Ronneberger, Fischer, and
  Brox]{Ronneberger2015UnetCN}
Olaf Ronneberger, Philipp Fischer, and Thomas Brox.
\newblock U-net: Convolutional networks for biomedical image segmentation.
\newblock In \emph{MICCAI}, 2015.

\bibitem[Shridhar et~al.(2022)Shridhar, Manuelli, and Fox]{shridhar2022cliport}
Mohit Shridhar, Lucas Manuelli, and Dieter Fox.
\newblock Cliport: What and where pathways for robotic manipulation.
\newblock In \emph{Conference on Robot Learning}, pages 894--906. PMLR, 2022.

\bibitem[Spencer et~al.(2020)Spencer, Choudhury, Barnes, Schmittle, Chiang,
  Ramadge, and Srinivasa]{Spencer-RSS-20}
Jonathan Spencer, Sanjiban Choudhury, Matt Barnes, Matthew Schmittle, Mung
  Chiang, Peter Ramadge, and Siddhartha Srinivasa.
\newblock {Learning from Interventions: Human-robot interaction as both
  explicit and implicit feedback}.
\newblock In \emph{Proceedings of Robotics: Science and Systems}, Corvalis,
  Oregon, USA, July 2020.
\newblock \doi{10.15607/RSS.2020.XVI.055}.

\bibitem[Tellex et~al.(2011)Tellex, Kollar, Dickerson, Walter, Banerjee,
  Teller, and Roy]{Tellex2011UnderstandingNL}
Stefanie Tellex, Thomas Kollar, Steven Dickerson, Matthew~R. Walter,
  Ashis~Gopal Banerjee, Seth~J. Teller, and Nicholas Roy.
\newblock Understanding natural language commands for robotic navigation and
  mobile manipulation.
\newblock In \emph{AAAI}, 2011.

\bibitem[Tellex et~al.(2014)Tellex, Knepper, Li, Rus, and
  Roy]{Tellex2014AskingFH}
Stefanie Tellex, Ross Knepper, Adrian Li, Daniela Rus, and Nicholas Roy.
\newblock Asking for help using inverse semantics.
\newblock In \emph{RSS}, 2014.

\bibitem[Tellex et~al.(2020)Tellex, Gopalan, Kress-Gazit, and
  Matuszek]{tellex2020-robots-that-use-language}
Stefanie Tellex, Nakul Gopalan, Hadas Kress-Gazit, and Cynthia Matuszek.
\newblock Robots that use language.
\newblock \emph{Annual Review of Control, Robotics, and Autonomous Systems},
  3\penalty0 (1):\penalty0 25--55, 2020.
\newblock \doi{10.1146/annurev-control-101119-071628}.
\newblock URL \url{https://doi.org/10.1146/annurev-control-101119-071628}.

\bibitem[Thomason et~al.(2022)Thomason, Shridhar, Bisk, Paxton, and
  Zettlemoyer]{thomason2022language}
Jesse Thomason, Mohit Shridhar, Yonatan Bisk, Chris Paxton, and Luke
  Zettlemoyer.
\newblock Language grounding with 3d objects.
\newblock In \emph{Conference on Robot Learning}, pages 1691--1701. PMLR, 2022.

\bibitem[Venkatesh et~al.(2021)Venkatesh, Upadrashta, and
  Amrutur]{venkatesh-iros2021}
Sagar~Gubbi Venkatesh, Raviteja Upadrashta, and Bharadwaj Amrutur.
\newblock Translating natural language instructions to computer programs for
  robot manipulation.
\newblock In \emph{2021 IEEE/RSJ International Conference on Intelligent Robots
  and Systems (IROS)}, pages 1919--1926, 2021.
\newblock \doi{10.1109/IROS51168.2021.9636342}.

\bibitem[Wallach and Allen(2008)]{Wallach2008MoralMT}
Wendell Wallach and Colin Allen.
\newblock Moral machines: Teaching robots right from wrong.
\newblock 2008.

\bibitem[Williams et~al.(2018)Williams, Gopalan, Rhee, and
  Tellex]{Williams2018LearningTP}
Edward~C Williams, Nakul Gopalan, Mine Rhee, and Stefanie Tellex.
\newblock Learning to parse natural language to grounded reward functions with
  weak supervision.
\newblock In \emph{ICRA}, 2018.

\bibitem[Xia et~al.(2013)Xia, L{\'e}onard, Kandaswamy, Blank, Whitcomb, and
  Kazanzides]{xia2013model}
Tian Xia, Simon L{\'e}onard, Isha Kandaswamy, Amy Blank, Louis~L Whitcomb, and
  Peter Kazanzides.
\newblock Model-based telerobotic control with virtual fixtures for satellite
  servicing tasks.
\newblock In \emph{2013 IEEE International Conference on Robotics and
  Automation}, pages 1479--1484. IEEE, 2013.

\end{thebibliography}

\clearpage
\appendix
\subsection{Planner}
\label{app:mpc}
We use STORM~\cite{bhardwaj2021storm} as the planner which computes an action leveraging sampling based optimization to optimize over costs. The base cost~$\costbasic$ for the 2D simulation robot contains the following terms:
\begin{align*}
  \mathcal{C}_\text{joint}(s_t) &= \begin{cases}
    ||s_t-s_{min}||_2 & \text{if }{s_t<s_{min}}\\
    ||s_{max} - s_{t}||_2 & \text{else if }{s_t > s_{max}}\\
    0 & \text{otherwise}
    \end{cases}\\
    \mathcal{C}_{collision}(q_t) &= Coll(q_t, o^h)
\end{align*}
where~$Coll(\cdot)$ checks for collisions between the robot position and the image~$o^h$ using a binary mask.

When using the planner on the Franka Panda robot, we use the cost terms described in~\cite{bhardwaj2021storm} with the following changes:
\begin{enumerate}
    \item We only check for environment collisions with the table. We don't check for collisions with objects and rely on~$\languagecost$ to ensure the ensure that the trajectory taken by the robot is not in collision.
    \item We add a cost to constrain the position along z-axis and 3D orientation of the gripper during execution to a default value that's close to the table.
\end{enumerate}
Across all instances of the planner, we use 500 particles and a horizon of 30 timesteps. 
\subsection{Hard environments}
\label{app:hard_envs}
\begin{figure}
    \centering
    \includegraphics[width=0.49\textwidth]{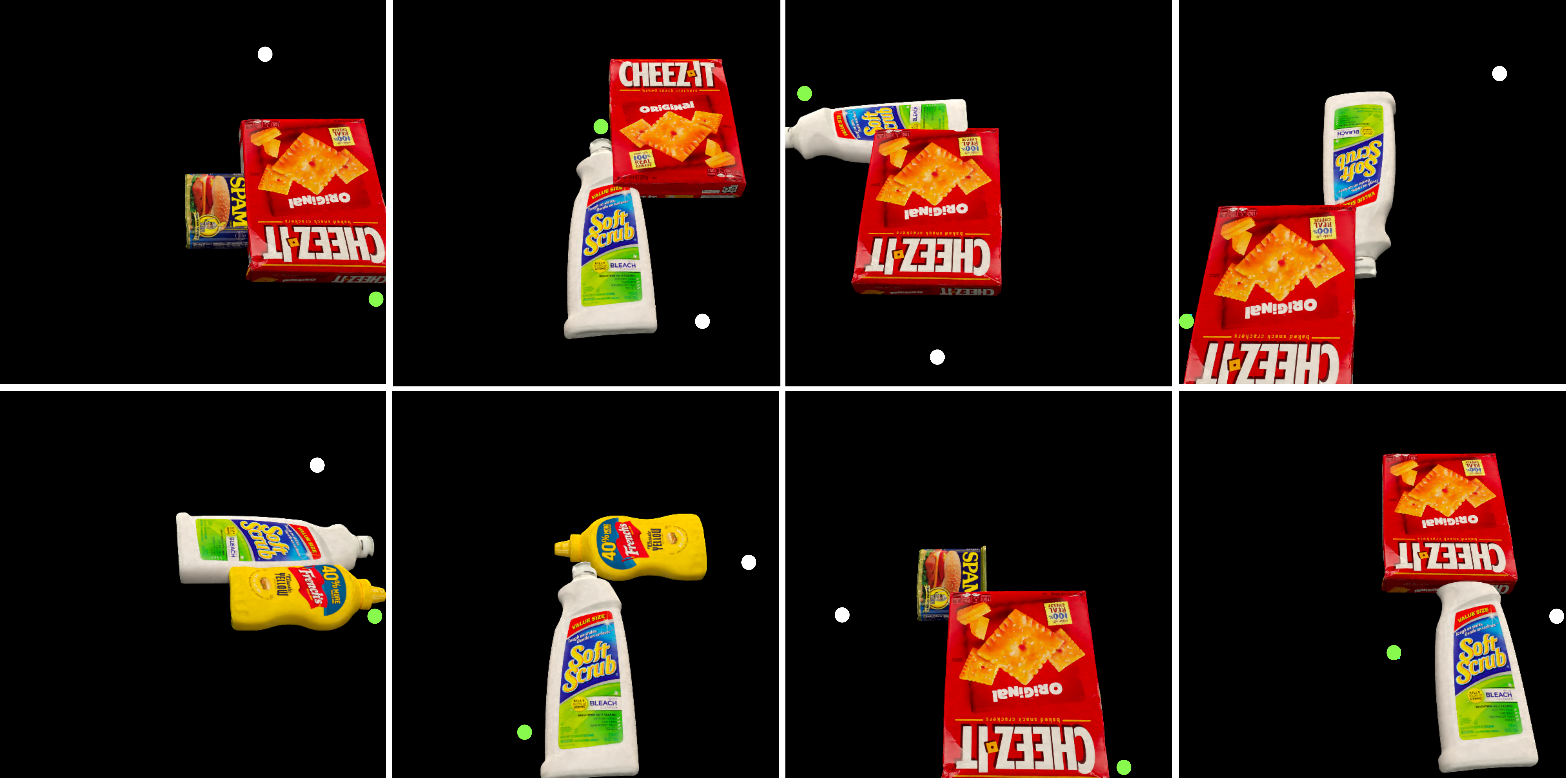}
    \caption{\textbf{Hard Environments:} Examples of some of the environments in the hard dataset. The White dot is the start location of the robot and the green dot is the goal location. The planner takes paths that lead it down to bad-local minima in these environments. }
    \label{fig:hard_envs}
\end{figure}

Some examples of environments where the planner fails can be seen in Figure~\ref{fig:hard_envs}. An MPC model minimizing the $\taskcost$ from the start location to goal gets stuck in hard to escape local-minima solutions. The robot is required to take several steps along a trajectory of increasing $\taskcost$ in order to reach a point starting where the robot can resume minimizing the specified $\taskcost$  to reach the true goal. It is these inflection points that the natural language feedback, $\langfeedback$, helps point the robot to.

\subsection{Convergence to Goal : Analysis}
\label{app:converge}

A natural question to ask is what can one do when the correction module itself fails and afterall, it is also a model not immune to failures. Here we understand when the correction module  fails. We group trajectories into easy medium and hard based on the length of the trajectories. It can be seen that corrections with longer trajectories are much worse than corrections with shorter trajectories. The main insight is that despite having a limited correcting ability one can still use it to make simple modifications at once or sequentially to correct behaviors.

\begin{figure}
    \centering
    \includegraphics[width=0.49\textwidth]{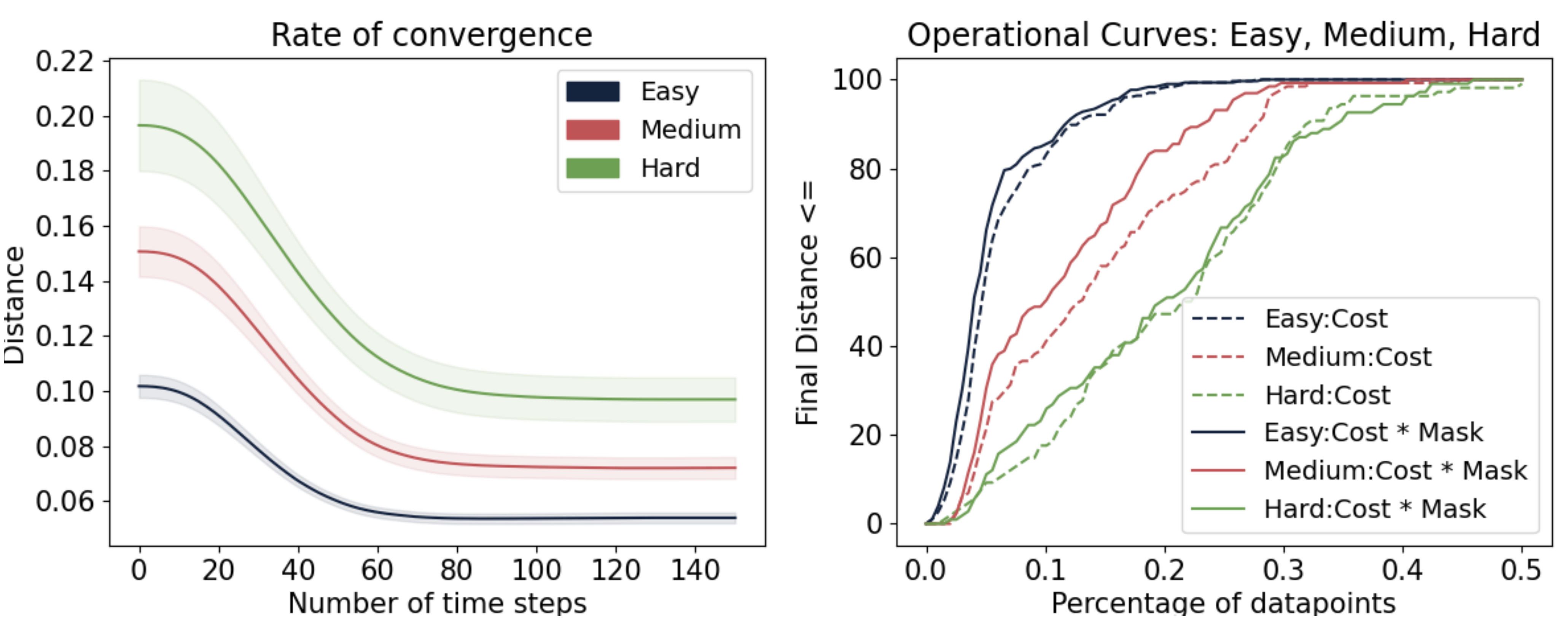}
    \caption{\textbf{Convergence to Goal Analysis:} a) As discussed in the results section the planner does better at short-horizon tasks as compared to long horizon tasks. The interesting aspect of the model is that even for long-horizon tasks, the first part of the trajectory does move in a direction where the goal is minimised for several steps. Even a model for $\languagecost$ with varying performance across short and long horizon corrections can still do well on introducing corrections that improve planner performance b) The advantage of generating a mask $\mask$ over the cost is most evident for medium to long trajectories.}
    \label{fig:convergence}
\end{figure}
\end{document}